# JOINT INTERPRETATION OF ON-BOARD VISION AND STATIC GPS CARTOGRAPHY FOR DETERMINATION OF CORRECT SPEED LIMIT

**Alexandre Bargeton, Fabien Moutarde, Fawzi Nashashibi and Anne-Sophie Puthon**
Robotics Lab (CAOR), Mines ParisTech
60 Bd St Michel, F-75006 Paris, FRANCE
Fabien.Moutarde@mines-paristech.fr

## ABSTRACT

We present here a first prototype of a "Speed Limit Support" Advance Driving Assistance System (ADAS) producing permanent reliable information on the current speed limit applicable to the vehicle. Such a module can be used either for information of the driver, or could even serve for automatic setting of the maximum speed of a smart Adaptive Cruise Control (ACC). Our system is based on a joint interpretation of cartographic information (for static reference information) with on-board vision, used for traffic sign detection and recognition (including supplementary sub-signs) and visual road lines localization (for detection of lane changes). The visual traffic sign detection part is quite robust (90% global correct detection and recognition for main speed signs, and 80% for exit-lane sub-signs detection). Our approach for joint interpretation with cartography is original, and logic-based rather than probability-based, which allows correct behaviour even in cases, which do happen, when both vision and cartography may provide the same erroneous information.

## INTRODUCTION AND RELATED WORK

Too many accidents are still provoked by excessive speed limit. It would therefore be interesting to offer drivers permanently updated information on current speed limit, so they have more chance to adapt their speed. Alternately, a more advanced very valuable feature would be a smart Adaptive Cruise Control (ACC) automatically adapting vehicle speed to current speed-limit. However, these kinds of functions can be really valuable and of practical use only if the produced speed-limit information is really reliable. Most current GPS navigators now include a function to inform the driver of the supposed current speed-limit, this information extracted from GPS cartographic data is neither always complete nor systematically up-to-date. Moreover, temporary speed limits for road works, and variable speed limits enforced by LED signs, are by definition not included in pre-defined digital cartographic data. Visual traffic sign recognition (TSR) can be quite robust, but it is unavoidable to miss some signs when occlusion by another vehicle (especially truck) occurs.

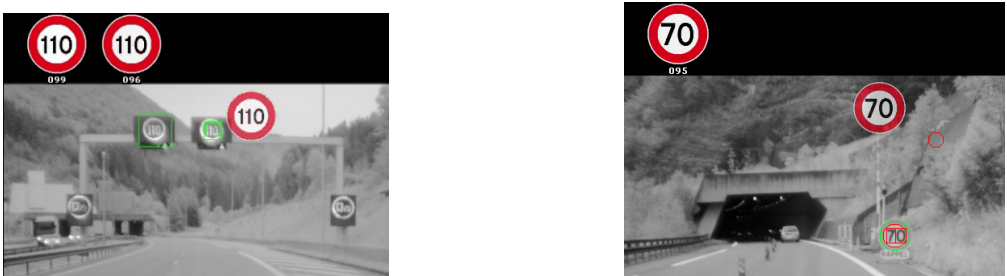

**Figure 1.** By nature, variable speed limits and roadwork temporary speed limits cannot be obtained from the static cartographic information.

Moreover, in order to determine the applicable speed limit by vision, it is also necessary to detect and recognize supplementary signs located below main signs and modifying their scope (class of vehicle, concerned lane, etc...), and to interpret correctly the vehicle itinerary in case of exit-lane-only speed signs.

## TRAFFIC SIGN RECOGNITION

Traffic Signs detection and Recognition (TSR) usually involves two main steps: 1/ detection of potential traffic signs in the image, based on the common shape/color design of sought traffic signs; 2/ classification of the selected regions of interest (ROI) for identifying the exact type of sign, or rejecting the ROI. Many TSR systems (e.g. [1], [2]) use color information to make detection step easier. But as noted and advocated in [3] and [4], using only shape information in grayscale improves robustness for operation in dark or night condition. This is what we do in our Speed Limit Support (SLS) prototype, already presented in [5] and [7], which relies on digit extraction and identification for recognition step (contrary to most TSR systems which use global recognition as in [1] [2] [3] [4] [6]). Our current TSR system is quite robust (~95% global correct detection on daytime) and fast for detecting and recognizing signs for *beginning* of speed-limit [5].

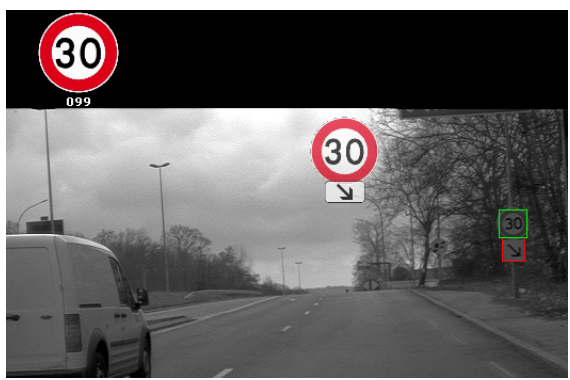

**Figure 2.** Illustration of detection and recognition of "arrow-type" supplementary sub-sign; note that proper handling of such sub-signs is essential for deducing actual speed limit, as they specify to which lane the speed limit of the traffic sign applies; other critical sub-sign include those for vehicle-specific or weather-specific speed-limits.

Our TSR system also includes a multi-country end-of-speed-limit signs recognition, as well as a detection and recognition of "exit-lane" supplementary sub-signs (see [7], and figure 2). To our knowledge, these last features are completely original.

## FUSION OF VISION WITH CARTOGRAPHY

Fusion of cartographic information with other sensors has begun to be experimented only rather recently. For the case of fusion of cartography with visual Traffic Sign Recognition, the very few published work are very recent, and all based on Belief Theory of Dempster-Schaefer [8][9][10]. These works therefore consider this fusion as a case of 2 complementary sensors with different reliabilities, from which the actual speed limit value can be deduced. Each hypothesis is assigned a mass of belief depending on the output of the 2 sensors, and what is retained is the speed limit which has a maximum plausibility. As we shall see below, although this approach may be pertinent in many circumstances, it cannot properly handle cases, which do happen, in which both sensor provide the same wrong answer.

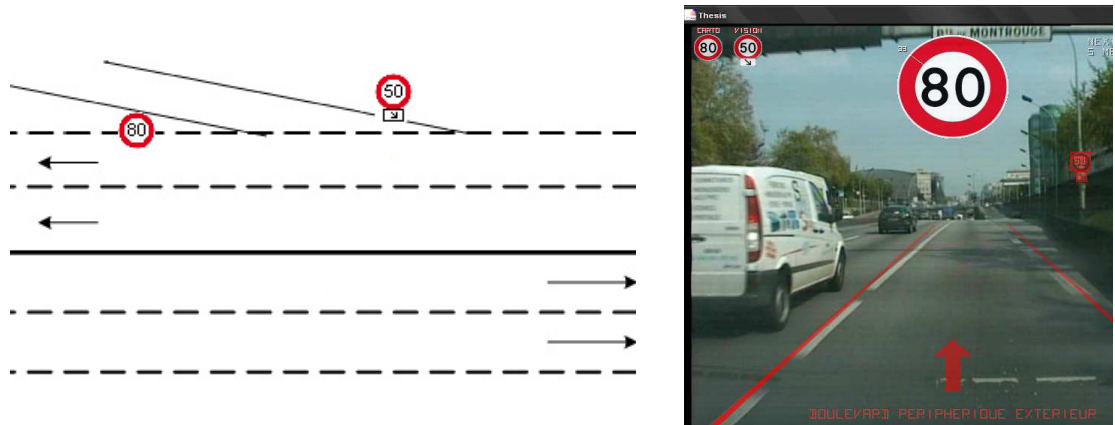

**Figure 3.** Left, an example in which both cartography and vision may provide the same wrong answer, when exiting; on the right, the decision between the 80 from cartography and the 50 from vision is more easily made by temporal logic than by some plausibility-based fusion of the 2 informations.

## PROPOSED LOGIC-BASED JOINT INTERPRETATION

Considering the quite common scenario illustrated on figure 3, it is clear that a probabilistic fusion scheme such as the Belief Theory is not very suitable, as the choice between 80 or 50 speed limits should rather be based on a logic interpretation of the sequence of observations and vehicle actions: the 50 speed limit concerning the exit-lane should be ignored unless the vehicle *later* takes this lane, which can be deduced reliably only by visual detection of lanes; then, in order to ignore the 80 speed sign one should implement a rule such as "when still on exit lane, ignore any *increase* of speed limit"; an interesting observation we have made during our experimental tests is that it may happen at this moment that both cartography (due to localization and/or map-matching error) and vision (due to the recognition on the left side of the 80 sign not concerning the exit-lane) provide the same erroneous speed limit (see figure 6), a case in which it is clear that any probabilistic decision between the two information is doomed to output an error.

We here propose a logic-based joint interpretation using as a third "sensor" a vision-based lane-change detector. The lane-detection uses Hough transform, and lane-changes are detected by time-evolution of lateral position of lanes.

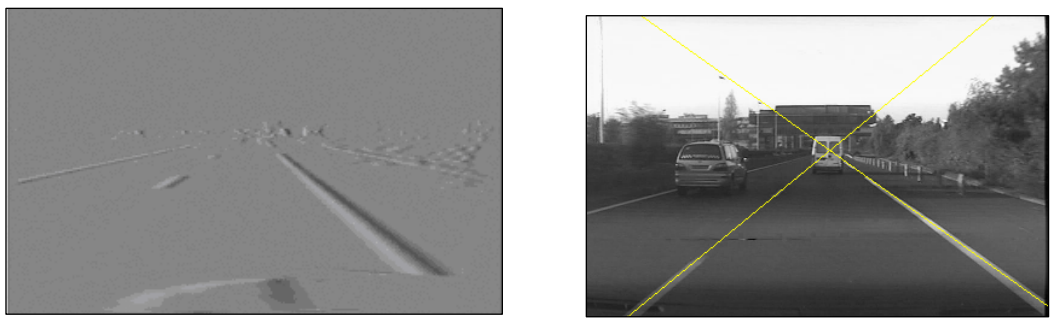

**Figure 4.** Lane-detection by Hough transform applied to gradient image; lane-changes are detected by analysis of lateral evolution of lanes in a memory of lane positions

The logic-based interpretation of vision+cartography+lane_changes can be described by a simple state-diagram, and a set of rules applying in the different states. Those are given in figure 5.

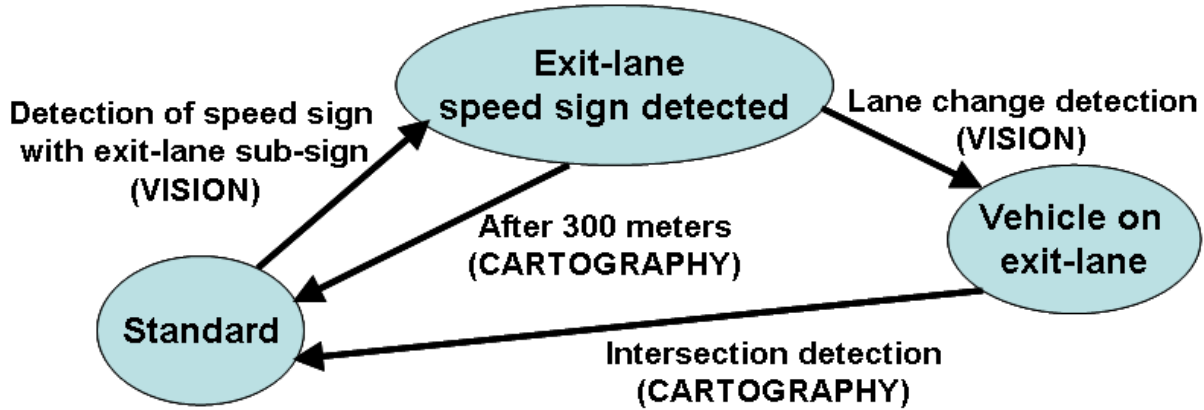

**State rules**
    Standard state → Validate vision limits *without* subsign
    « Exit-lane sign detected »   → Validate vision limits *without* subsign
    Exit-lane state → Validate any *decreasing* limits

**General rule**
IF currently validated limit is too old AND no risk of ambiguity in cartography (no very close road with other speed limit) THEN adopt current cartographic limit

**Figure 5.** Our proposed logic joint interpretation of cartography, visual TSR, and visual lane change detection: on top, the state-diagram, and below the set of rules.

This proposed system was successfully tested in real-time on-board our test vehicle on several typical potentially problematic situations such as that illustrated on figures 3 and 6.

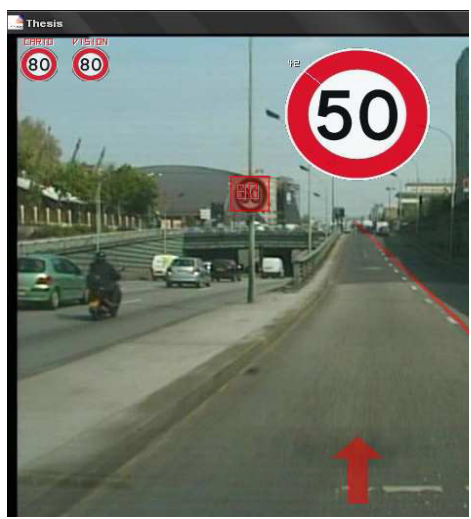

**Figure 6.** A very interesting case, in which our system correctly determines the 50 limit, even though cartography *and* vision provide the same wrong answer!

# DISCUSSION AND PERSPECTIVES

Our proposed system will be tested on more situations and longer sequences, in particular to tune the main parameters of our logic. We shall also investigate if more states and rules are required to handle other common problematic cases such as when a traffic sign is planted between two adjacent roads. Current and future works also include the addition of proper interpretation of other most common supplementary sub-signs.